\documentclass[letterpaper, 10 pt, conference]{ieeeconf}
\IEEEoverridecommandlockouts                              

\usepackage{amsthm}
\usepackage{enumerate}
\usepackage{amsfonts} 
\usepackage{amsmath} 
\usepackage{algorithm}
\usepackage{booktabs}
\usepackage{mathtools}

\usepackage{algpseudocode}
\usepackage{amssymb}
\usepackage{hyperref}
\usepackage{multicol}
\usepackage{physics}
\usepackage{xcolor}
\usepackage{url}
\usepackage{lipsum,adjustbox}
\usepackage{tikz}
\usepackage[font=small,labelfont=bf]{caption}
\usepackage{subcaption}
\usepackage{hhline}
\usepackage{multirow}
\let\labelindent\relax
\usepackage[shortlabels]{enumitem}
\renewcommand{\theenumi}{(\roman{enumi})}

\usepackage{cite}

\theoremstyle{theorem}
\newtheorem{Remark}{Remark}
\theoremstyle{definition}

\DeclareMathOperator*{\argmax}{arg\,max}

\title{Failure Prediction from  Limited Hardware Demonstrations}

\author{
    Anjali Parashar, 
    Kunal Garg, 
    Joseph Zhang,
    and Chuchu Fan
\thanks{The authors are with the Department of Aeronautics and Astronautics, Massachusetts Institute of Technology, USA. Email: {\tt\small \{anjalip,kgarg,jzha,chuchu\}@mit.edu}}
   }

\begin{document}
\maketitle

\begin{abstract}
Prediction of failures in real-world robotic systems either requires accurate model information or extensive testing. Partial knowledge of the system model makes simulation-based failure prediction unreliable. Moreover, obtaining such demonstrations is expensive, and could potentially be risky for the robotic system to repeatedly fail during data collection. This work presents a novel three-step methodology for discovering failures that occur in the true system by using a combination of a limited number of demonstrations from the true system and the failure information processed through sampling-based testing of a model dynamical system. Given a limited budget $N$ of demonstrations from true system and a model dynamics (with potentially large modeling errors), the proposed methodology comprises of a) exhaustive simulations for discovering algorithmic failures using the model dynamics; b) design of initial $N_1$ demonstrations of the true system using Bayesian inference to learn a Gaussian process regression (GPR)-based failure predictor; and c) iterative $N - N_1$ demonstrations of the true system for updating the failure predictor. To illustrate the efficacy of the proposed methodology, we consider: a) the failure discovery for the task of pushing a T block to a fixed target region with UR3E collaborative robot arm using a diffusion policy; and b) the failure discovery for an F1-Tenth racing car tracking a given raceline under an LQR control policy.
\end{abstract}

\begin{figure*}[t]
  \includegraphics[height=3.3 in]{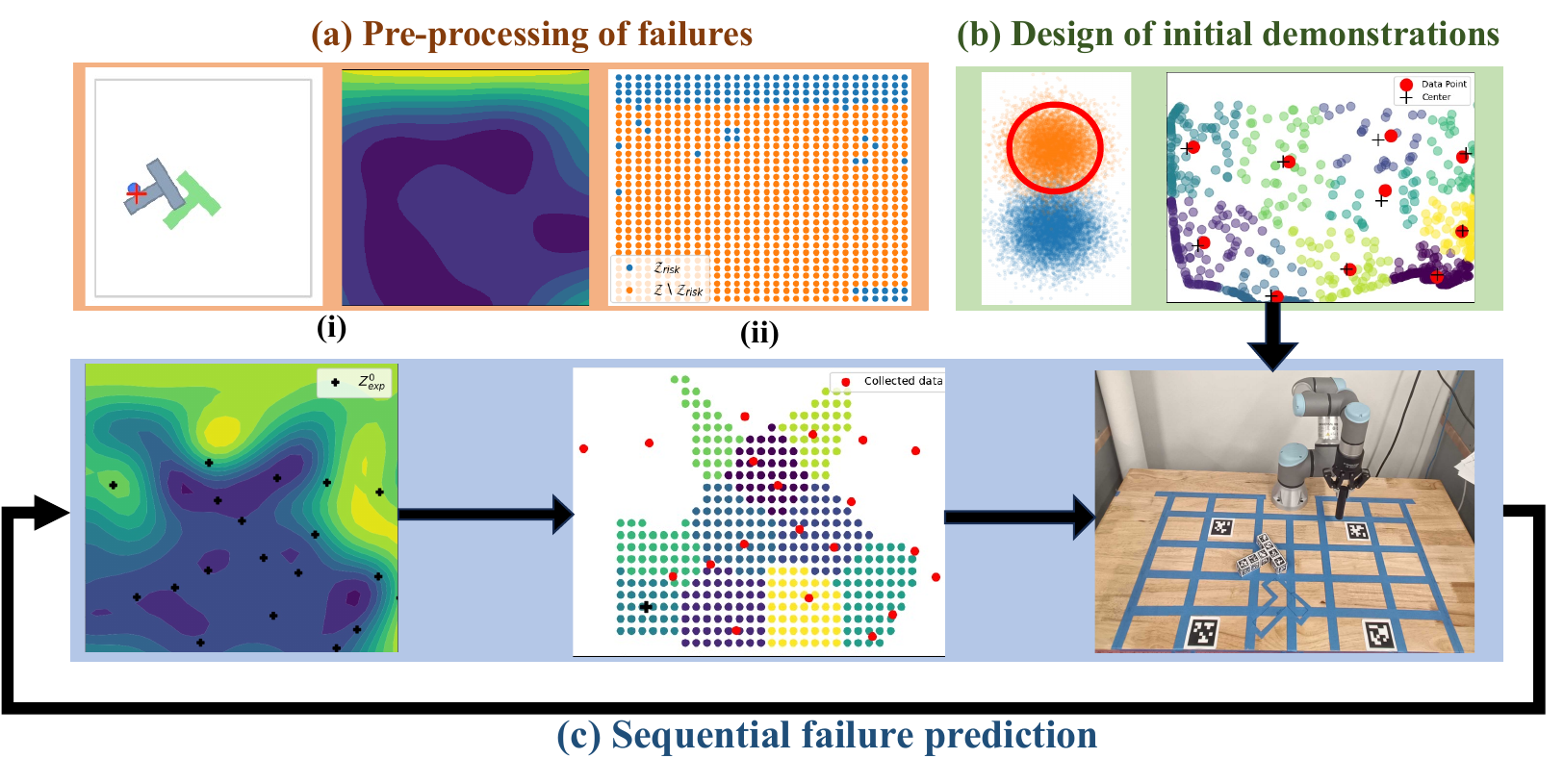}
  \caption{The proposed methodology constitutes a) discovering failures using model information; b) design of initial demonstrations to learn true system failures using Bayesian inference; and c) sequential demonstration from low predicted-risk regions for GPR-based risk prediction update.}
  \label{fig:method}
\end{figure*}

\section{Introduction}
Testing and model validation are essential tools for ensuring the safety of autonomous systems before deployment \cite{dreossi2015efficient,esposito2005adaptive,corso2020scalable,corso2019adaptive,sinha2020neural}. These techniques primarily pertain to discovering failures using available model information to study the limitations of decision-making pipelines and their subsequent improvement~\cite{sinha2020neural,yangSynthesisguidedAdversarialScenario2021,dontiAdversariallyRobustLearning2021,cavorsi2023exploiting,ghaiGeneratingAdversarialDisturbances2021,xuSafeBenchBenchmarkingPlatform2022}. Based on the extent of information available about the model, 
various model-based tools for testing and falsification tools have been proposed in the literature~\cite{annpureddySTaLiRoToolTemporal2011b,chouUsingControlSynthesis2018,delecki2023model,dawson2023a,dawsonbeyond,10669181}. The model-based tools have the advantage of being faster \cite{delecki2023model,dawson2023a}, and with differentiable system models, gradient-based optimization strategies can be used for falsification, which have shown to be more efficient than black-box methods \cite{dawson2022robust,ma2019sampling,ma2021there}. They also allow exploitation of model-specific information to expedite failure discovery.  

Recently, some works have considered simulation-based falsification and testing \cite{dawson2023a,xu2023diffscene}. These methods minimize the dependency on hardware for testing, which can be extremely time-consuming and require extensive utilization of resources \cite{koopman2017autonomous}. These methods can be further divided into three categories based on the underlying framework used. The first set of methods uses adversarial optimization to return a single failure per search~\cite{dawson2022robust,hanselmannKINGGeneratingSafetyCritical2022a,wong2018provable}. Some of these methods suffer from challenges such as getting stuck in local minima, due to their dependency on gradient-based optimization~\cite{du2017gradient}. The second class of methods utilizes learning-based approaches, with recent advances in generative modeling, to generate failure scenarios using information learned from the training dataset, and requires a sufficient amount of training data~\cite{kundu2024data,xu2023diffscene}, which can be challenging, since failures in autonomous systems are often low probability events or corner cases~\cite{delecki2023model,chou2018using,sunCornerCaseGeneration2021}. The third set of methods leverage probabilistic inference and sampling-based techniques, which return multiple failures per search~\cite{zhao2017accelerated,okellyScalableEndtoEndAutonomous2018a}, however, also suffer from issues of slow convergence and uneven search space exploration. Recently, some works have explored combinations of these fundamental techniques to overcome the limitations of each~\cite{delecki2023model, dawson2023a,10669181}. 

Most of these approaches assume that the simulation dynamics, which we refer to as the \textit{model dynamics}, and testing environment represent the realistic testing conditions adequately. However, this can be misleading, especially if the \textit{true system} and its environmental interactions are complex and cannot be adequately captured using appropriate simulation models~\cite{10669181}. Additionally, there are uncertainties in state estimation and dynamics, that affect the performance of true systems~\cite{ren2023adaptsim}. Collectively, these issues lead to failure modes that remain undiscovered, despite exhaustive simulation testing, due to the inadequacy of the simulation models in capturing wide-ranging practical phenomenon that affect the true system. 
This can impact the performance of autonomous systems, which is concerning from the perspective of safety. It is possible that the discovered failure modes in simulation do not reflect the true severity of real failures and that a scenario reported as safe in simulation may correspond to failure in real-time. 
This study investigates the discovery of failures that are hard to find just from the model dynamics due to sim-to-real gap and a limited budget for testing the true system. We explore the utility of Gaussian Process Regression (GPR) to extract failure information from true systems with limited demonstrations~\cite{iwazaki2024failure,nguyen2008local}, in addition to conventional exhaustive simulation-based testing. We leverage random sampling for testing the model dynamics and adopt a sampling-based perspective for failure discovery.  

In this paper, we investigate the sim-to-real gap from the perspective of falsification, using existing testing pipelines for simulation, while working with limited data from the true system to enable better prediction of failures (Section. \ref{sec:Problem}). Our proposed methodology consists of three steps (Section. \ref{sec:methodology}), namely, data collection of failure information from model dynamics, followed by an initial set of demonstrations on the true system, which is used to train an initial failure prediction model using GPR. This is followed by a sequential demonstration of the true system to uncover failures that cannot be captured by the simulations and subsequent improvement of the failure prediction model. In this step, the demonstrations are chosen to either validate that the regions defined as `not fail' by the then-learned failure predictor are truly safe or correspond to an undiscovered failure. Fig.~\ref{fig:method} summarizes the proposed methodology.

We validate our approach on two robotic examples, namely, pushing a T-block in a designated region~\cite{florence2022implicit} with a Diffusion policy using the UR3E collaborative robot end-effector~\cite{chi2023diffusionpolicy}, and reference path-tracking with an LQR Speed+Steering controller~\cite{sakai2018pythonrobotics} using the F1-Tenth autonomous racing platform~\cite{o2019f1}. For both problems, the policies are designed using simulation environments which represent simplified estimations of the true dynamic systems, and are subsequently implemented on the true dynamic systems. This leads to failure modes that are not observed in simulation, which we aim to discover using our methodology. 
Our methodology is able to discover failures with an accuracy of $89\%$ and $100\%$ as compared to $11\%$ and $36\%$ by a purely simulation-based failure predictor, in the first and the second example, respectively.

\section{Problem Formulation}\label{sec:Problem}
We consider two sets of dynamics in this paper: model (known to the user) and true (unknown to the user). \textit{True dynamics} corresponds to the actual dynamics of the agent, which is unknown whereas \textit{model dynamics}, or simply, model, corresponds to the estimate of the true dynamics and is described as:
\begin{equation}\label{eq:system}
    \begin{aligned}
        x_{t+1} & = f(x_{t},\pi(y_{t},z)) + \epsilon_1, \quad \quad \quad 
        y_{t} =  Cx_{t} + \epsilon_2,
    \end{aligned}
\end{equation}
where $f:\mathbb R^n\times\mathbb R^m\to \mathbb R^n$ and $C\in \mathbb R^{l\times n}$,
with state $x_t \in \mathbb{R}^{n}$ at time $t\in \mathbb R$ and policy $\pi: \mathbb{R}^{l} \times \mathbb R^d\to \mathcal{U} \subseteq \mathbb{R}^{m}$ which outputs actions based on environmental variables $z \in \mathcal{Z} \subset \mathbb{R}^{d}$ and system output $y_t\in \mathbb R^l$, where $\epsilon_1$ and $\epsilon_2$ are disturbances in dynamics and state estimation, respectively. The environment variable $z$ represents independent variables, such as initial conditions $x_0$ of the system  along with the environmental information exogenous to the system. We assume that the disturbances come from zero-mean Gaussian distributions given by $\epsilon_1 \sim \mathcal{N}(\boldsymbol{0},\Sigma_1)$ and $\epsilon_2 \sim \mathcal{N}(\boldsymbol{0},\Sigma_2)$, where the covariance matrices $\Sigma_1 \in \mathbb{R}^{n \times n},\Sigma_2 \in \mathbb{R}^{l \times l}$ of the distributions are defined using scalars $\sigma_1,\sigma_2 \geq 0$ as $\Sigma_i = \sigma_i \boldsymbol{I}$, for $i=1,2$, where $\boldsymbol{I}$ is an identity matrix of the appropriate size. For a given $z$, we denote a trajectory rollout of the model \eqref{eq:system} under a given $(\sigma_1, \sigma_2)$ as $X_{(z|\sigma_1, \sigma_2)}=(x_i)_{i=0}^{T}$ and of the true system as $X^*_z$.\footnote{In what follows, we suppress the explicit dependence on $\sigma_1, \sigma_2$ for the sake of brevity.} 

For the purpose of failure prediction, we consider a user-defined risk function $R:\mathbb R^d \to \mathbb{R}$ where $R(z) = R(z, X_z)$ denotes the risk corresponding to the trajectory rollout $X_z$ for a given environment variable $z$. Based on this risk function, we define failure of the system for a corresponding $z$ when the risk $R(z)$ exceeds a user-defined threshold $R_{\text{th}} \in \mathbb R$. The falsification of the closed-loop system, or plainly, failure discovery problem can be mathematically formulated as discovering the set 
    $\mathcal{Z}_{\text{fail}}^*\coloneqq\{z\; |\; R(z, X^*_z)>R_{\text{th}}\}$.
We aim to solve this under the constraint that we can query the true system only a limited times $N>0$ to obtain $N$ trajectory rollouts $\{X^*_{z_i}\}_{i=1}^N$ for $z_i\in \mathcal Z$. We present a three-step methodology to discover failures occurring in the true system by using a combination of a minimal number of demonstrations $\{X^*_z\}_N$ and the failure information from the model dynamics (denoted as $\mathcal Z_\text{fail}(f)$) obtained through sampling-based falsification (see Figure \ref{fig:method}). In brief, the proposed methodology constitutes a) exhaustive simulations for discovering algorithmic failures using the model dynamics; b) design of initial $N_1 < N$ demonstrations from the true system using Bayesian inference to learn a Gaussian process regression (GPR)-based failure predictor; and c) iterative $N - N_1$ demonstrations of the true system for updating the failure predictor, where $N_2=N - N_1$.

\section{Methodology}\label{sec:methodology}
\begin{figure}
    \centering
    \includegraphics[height=1.5 in]{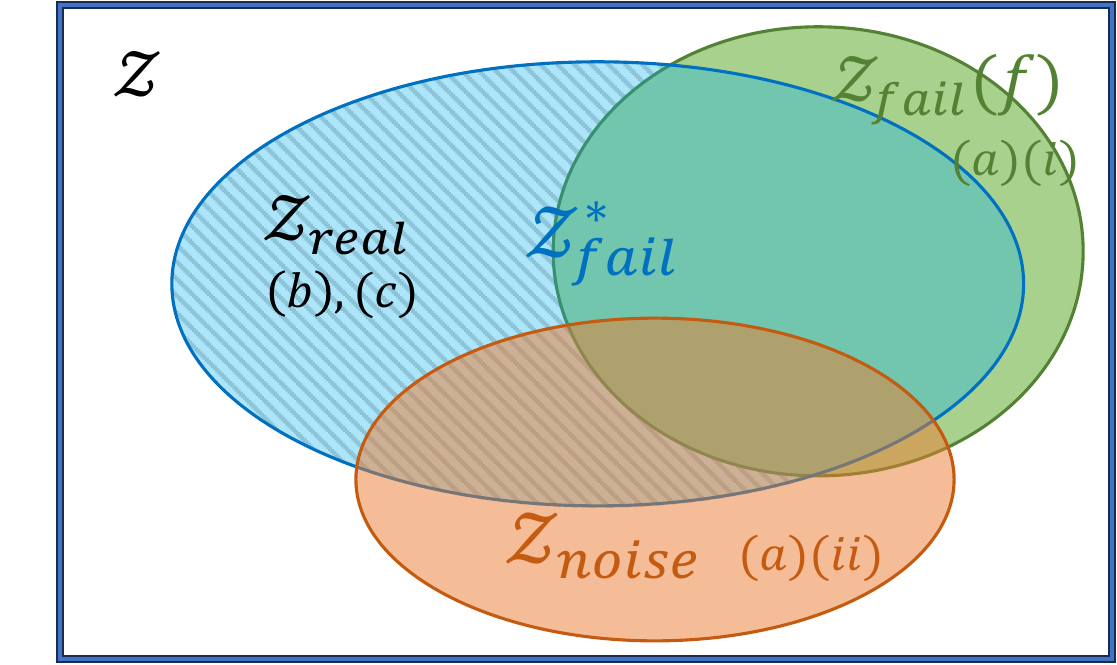}
    \caption{Venn Diagram showing different sets defined in Section.~\ref{sec:methodology}. Here, $\mathcal{Z}_{\text{fail}}(f)$ and $\mathcal{Z}_{\text{noise}}$ denote the set of failures that can be discovered using model dynamics (Fig.~\ref{fig:method}-a. (i) and (ii) respectively), and  $\mathcal{Z}_{\text{real}}$ is estimated using demonstrations from the true system (Fig.~\ref{fig:method}-b. and c.)}
    \vspace{-15pt}
    \label{fig:Venn_diagram}
\end{figure}

We assume that the model \eqref{eq:system} can capture certain failures that could occur with the true system, i.e., $\mathcal{Z}_{\text{fail}}(f) \cap \mathcal{Z}_{\text{fail}}^* \neq \emptyset$. Based on this, we obtain that $\mathcal{Z}_{\text{fail}}^* \subseteq \mathcal{Z}_{\text{fail}}(f) \cup \mathcal Z_{\text{real}}$, i.e, failures of the true system are a combination of algorithmic failures on the model system $\mathcal{Z}_{\text{fail}}(f)$ and failures due to the mismatch between model dynamics and actual dynamics, disturbances and other potentially unknown reasons, cumulatively represented by $\mathcal Z_{\text{real}}$. We say that the set $\mathcal Z_{\text{fail}}(f)$ captures algorithmic failures as we assume that the policy $\pi$ in \eqref{eq:system} is designed for the model $f$ but still leads to failures. 
The first step of our methodology focuses on discovering failures that can be obtained using the model information. 

\subsection{Pre-processing of failures: utilizing model information}\label{sec:Flowgmm}
We define the set of the environment variables for the algorithmic failures of the model dynamics as:
\begin{equation}\label{eq:Z_fail}
        \mathcal Z_{\text{fail}} (f) \coloneqq \{z\; |\; R(z, X_{(z|\sigma_1=0,\sigma_2=0)})>R_{\text{th}} \},
\end{equation}
This set can be obtained through extensive simulations using the model information. Next, we aim to capture the failures due to the mismatch between model dynamics and true dynamics, disturbances, and potentially other unknown reasons. 
For this, we sample $\sigma_1, \sigma_2$ from a bounded region given by $\mathcal{B} \coloneqq [\sigma_1^{\text{min}},\sigma_1^{\text{max}}]\times [\sigma_2^{\text{min}},\sigma_2^{\text{max}}]$, and using the risk $R(z,X_z)$ corresponding to each disturbance, we collect values of $z$ for which a failure is observed across all disturbances.  Here, the maximum values of the variance $\sigma_1^{\text{max}}, \sigma_2^{\text{max}}$ are chosen high enough to appropriately represent the scale and magnitude of disturbances experienced in the true system, based on expert knowledge. 
We define $\mathcal Z_{\text{noise}}$ as the set of $z$ for which the model system fails due to disturbances as:
\begin{equation}\label{eq:Z_risk}
        \mathcal Z_{\text{noise}} \coloneqq \{z\; |\; R(z, X_{(z|\sigma_1,\sigma_2)})>R_{\text{th}} \; \forall \sigma_1,\sigma_2 \in \mathcal{B}\setminus \{[0, 0]\}\},
\end{equation}
so that $\mathcal Z^*\cap \mathcal Z_{\text{noise}} \neq \emptyset$. 
The set $\mathcal Z_{\text{risk}} \coloneqq \mathcal Z_{\text{noise}} \cup \mathcal Z_{\text{fail}}(f)$ captures all possible failures that can be discovered using model dynamics \eqref{eq:system}. Fig.~\ref{fig:Venn_diagram} illustrates these various failure sets and their inter-relationships. The next step is to discover failures that the model system cannot capture through sampling $z$ from the region $\mathcal Z_{\text{real}}$ and obtaining demonstrations from true dynamics. Since $\mathcal Z_\text{real}$ is not known, we obtain these samples from the region $\mathcal{Z} \backslash \mathcal Z_{\text{risk}}$ as discussed next. 

\subsection{Sampling from sensitive regions: Design of experiments}\label{sec:BayesInf}
Since we have a limited budget on the number of demonstrations we can obtain from the true dynamics, we choose to collect the demonstrations that maximize the state-space coverage. For a given $z$, define a coverage function $C:\mathcal Z\to \mathbb R$ where $C(z) = C(z, X_z)$ is a monotonically increasing function of the explored state-space along the trajectory $X_z$ (see Section \ref{sec:validations} for examples of the coverage functions used in practice). With this, we aim to sample $z\in \mathcal{Z} \backslash \mathcal Z_{\text{risk}}$ from a region corresponding to high coverage, given by:
\begin{equation}\label{eq:z_posterior}
    z \sim \{ z|C(z, X_z)>C_{\text{th}}, z \in \mathcal{Z} \backslash \mathcal Z_{\text{risk}}\},
\end{equation}
where $C_{\text{th}} > 0$ is a user-defined coverage threshold. 
Sampling directly from this set is intractable, and so, we utilize a Bayesian inference framework here~\cite{zhou2022rocus,10669181}. 
To sample exclusively from $\mathcal{Z} \backslash \mathcal Z_{\text{risk}}$, we learn the decision boundary that distinguishes $\mathcal Z_{\text{risk}}$ from the remaining search space by performing supervised binary classification. For this, we use a specific technique known as Flow-GMM \cite{izmailov2020semi}, which allows us to learn an invertible mapping from search-space $\mathcal{Z}$ to a latent-space $\mathcal{W} \in \mathbb{R}^{d}$, given by $g^{-1}_{\theta}: \mathcal Z\to \mathcal W$ using the Normalizing Flows framework~\cite{dinh2016density}. Using Flow-GMM, we learn isotropic Gaussian latent distributions $\tilde{p}_1,\tilde{p}_2$ corresponding to the regions $\mathcal{Z}_{\text{risk}}$ and $\mathcal{Z} \backslash \mathcal Z_{\text{risk}}$ which can be mathematically expressed using mean $\mu^g_i$ and covariance $\Sigma^g_i$ as $\tilde{p}_i = \mathcal{N}(\mu^g_i,\Sigma^g_i)$ for $i=1,2$. 
We can rewrite sampling from the set in~\eqref{eq:z_posterior} to sampling from a distribution in the latent space $\mathcal{W}$ using the learned mapping $g_{\theta}$ as:
\begin{equation}\label{eq:w_posterior}
        w \sim \tilde{p}_2(w|C(g_{\theta}(w), X_{g_{\theta}(w)})>C_{\text{th}})  
\end{equation}

We utilize exponential modelling for expressing the likelihood $p(C>C_{\text{th}}|g_{\theta}(w))$, adopted from \cite{zhou2022rocus} and $\tilde{p}_2$ as the prior, to construct the posterior distribution for sampling. Using Bayes rule,~\eqref{eq:w_posterior} can be simplified as:
\begin{equation}\label{eq:w_posterior_2}
\begin{aligned}
    w & \sim  \tilde{p}_2(w|C(g_{\theta}(w), X_{g_{\theta}(w)})>C_{\text{th}}) \\
    & \propto \exp \big ( -[C_{\text{th}} - C(g_{\theta}(w), X_{g_{\theta}(w)})]_{+} \big) \tilde{p}_2(w).
\end{aligned}
\end{equation}
Here $[]_{+}$ represents the ReLU operator, and ensures that each $w$ corresponding to $C(g_{\theta}(w), X_{g_{\theta}(w)})\geq C_{\text{th}}$ is prioritized equally. 
We use Metropolis-Hashtings algorithm \cite{robert2004metropolis} to sample $w$ from the constructed posterior distribution, which can also be replaced by another sampling algorithm of choice. 


While we are sampling from the posterior~\eqref{eq:w_posterior}, the generated iterates must satisfy $g_{\theta}(w) \in \mathcal{Z} \backslash \mathcal{Z}_{\text{risk}}$. This is done by using a projection operator $\mathbb{P}[w]$ to ensure that the generated samples lie within a convex set centered at the mean $\mu_2^g$, so that the projected sample has a high probability of being in the set $\mathcal{Z} \backslash \mathcal{Z}_{\text{risk}}$. Since $\tilde{p}_2(w)$ is isotropic, we chose $\boldsymbol{P}[w] = c + r\frac{w-c}{\lVert w-c \rVert_2}$ with $c=\mu_2^g$, and $r$ as a user-defined variable that can be decreased to make the sampling more conservative or vice versa. It can be easily verified that for a point $w$ such that $\lVert w -c \rVert_2 > r$, the projected point $\tilde{w}=\boldsymbol{P}[w]$ satisfies $\lvert w -c \rvert_2 \leq r$, therefore the mapping of the projected sample $g_{\theta}(\tilde{w})$ has a high probability of lying in the region $\mathcal{Z} \backslash \mathcal{Z}_{\text{risk}}$ for an appropriately chosen $r$. Fig.~\ref{fig:method} (ii)-(iii) show the latent distribution learned using Flow-GMM corresponding to $\mathcal{Z}_{\text{risk}}$ and $\mathcal{Z} \backslash \mathcal{Z}_{\text{risk}}$, and the boundary of the constructed set $\mathcal{P}$ in red for the Push-T task discussed in Section.~\ref{sec:pusht}

The pipeline discussed so far generates a collection of samples $Z_\text{cov}=\{z \in \mathcal{Z} \backslash \mathcal Z_{\text{risk}} \;|\;C(z,X_z)>C_{\text{th}}\}$. Once we have generated the samples, we choose $N_1$ candidate values of $z$ distributed uniformly across the search-space. This is achieved by dividing $Z_\text{cov}$ into $N_1$ clusters using $k$-means clustering, and choosing the points in $Z_\text{cov}$ closest to the geometric centers of the generated clusters for demonstrations.  This allow us to collect $N_1$ risk values given by $R_1=\{R(z_j, X_{z_j}^*)\}$ for the data points $Z_1=\{z_j\}$. We also obtain $M$ data points $Z_2=\{z_i\}$ from the region $\mathcal Z_{\text{risk}}$ (Section.~\ref{sec:Flowgmm}), with the corresponding risk values given by $R_2=\{R(z_i, X_{z_i})\}$ using model dynamics $f$. Define $\mathcal{D}_1 = [Z_1, R_1]$ and $\mathcal{D}_2=[Z_2, R_2]$ as the dataset of demonstrations obtained from true and model systems, respectively. We use these dateset to train a GP model $\phi_\theta:\mathcal{Z} \to \mathbb{R}$ to predict risk $\hat{R}=\phi_\theta(z)$ for a given $z$, as illustrated in the next section where $\theta$ denotes the model parameters. 

\subsection{Sequential failure prediction and training using GPR}\label{sec:GPR}

Motivated by the success of GPR in learning from limited demonstrations \cite{nguyen2008local}, we use GPR as the backbone of the risk prediction pipeline. 
Using the dataset $\mathcal{D}_f = \mathcal{D}_1 \cup \mathcal{D}_2$, we construct the marginal log likelihood $\log p_{\phi_\theta}(R_f|Z_{\text{f}})$ for learning the model $\phi_\theta$ using a sum of the marginal log likelihoods from both sources of data as:
\begin{equation}\label{eq:weighted-mll}
    \log p_{\phi_\theta}(R_f|Z_{\text{f}}) = \log p_{\phi_\theta}(R_1|Z_1) + \log p_{\phi_\theta}(R_2|Z_2), 
\end{equation}
where $Z_f=[Z_1,Z_2]$ and $R_f=[R_1,R_2]$. 
The training objective is the maximization of the marginal log likelihood in~\eqref{eq:weighted-mll} with $\phi_\theta$ as the decision variable:
\begin{equation}\label{eq:argmax}
    \phi_{\theta_0} = \argmax_{\theta} \log p_{\phi_\theta}(R_f|Z_f).
\end{equation}
We train the risk predictor model $N_2$ times by a sequence of $N_2$ demonstrations on the true system with sequential optimization of $\theta$ by solving~\eqref{eq:argmax} and generation of a new data-point $z$ for the next demonstration. In particular, at each step $k$, we first divide the region $\{z \; |\; \phi_{\theta}(z) < R_{\text{th}}\}$ which is predicted `not fail' by the learned model into $N_2$ clusters using $k$-means clustering. We choose $z_k$ from the set of geometric means $c_i$ of the clusters as the point which maximizes the distance from the previously chosen points:
\begin{equation}
    z_k = \argmax_{z \in \mathcal{C}_k} \min_{q \in [Z_1,Z_{k-1}]} \lVert z-q \rVert^2_2 \quad k=1,\dots,N/2 ,
\end{equation}
where $\mathcal{C}_k = [c_1,\dots,c_{N/2}]$ and $Z_{k-1} = [z_1,\dots,z_{k-1}]$. Each step of demonstration is followed by training of the GPR model $\phi_{\theta}$ with the dataset updated with $z_k$:
\begin{equation}\label{eq:bilevel}
    \phi_{\theta_k}= \argmax_{\theta}  \log p(\hat{R}|\phi_{\theta},[Z_{\text{f}},(z)_{i=1}^{k}]) 
\end{equation}
Since the number of data points is limited, training the GPR model $N_2$ times is not challenging in practice. For a larger dataset, updating the pre-trained model instead of re-training can be computationally more efficient. 

\begin{Remark}
The proposed methodology has the advantage of being highly modular. For example, when working with systems with small dimensional environmental variables (e.g., $z\in \mathbb R^2$ or $z\in \mathbb R^3$)
a na\"ive random testing on the model dynamics is sufficient for generating the samples corresponding to $\mathcal{Z}_{\text{risk}}$. 
This can be substituted with a Bayesian inference pipeline and alternative sampling methods for testing in higher dimensional search space. Additionally, the methodology uses $k$-means clustering and selection of geometric centers for demonstrations of the true system. This is one of the several possible ways in which $N$ data points for demonstration can be selected. 
\end{Remark}

\begin{figure}
\centering
\begin{subfigure}{0.4\linewidth}
    \includegraphics[width=\textwidth, trim={0 2.1cm 0 0},clip]{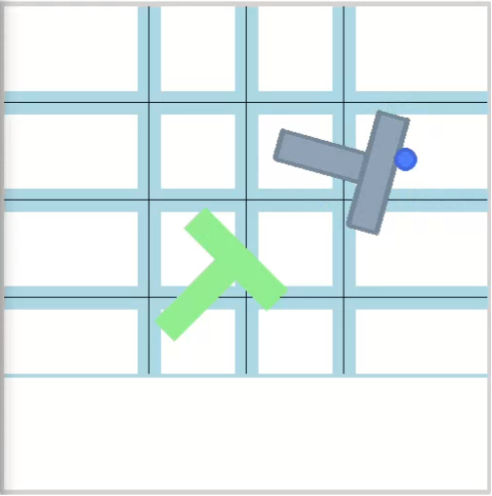}
    \caption{Model system}
    \label{subfig:Pusht_Sim}
\end{subfigure}
\hspace{20pt}
\begin{subfigure}{0.4\linewidth}
    \includegraphics[width=\textwidth]{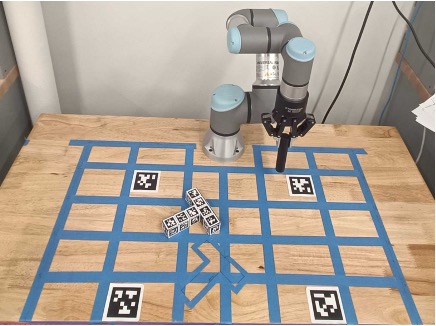}
    \caption{True system}
    \label{subfig:Pusht_real}
\end{subfigure}
\centering
\begin{subfigure}{0.4\linewidth}
    \includegraphics[width=\textwidth]{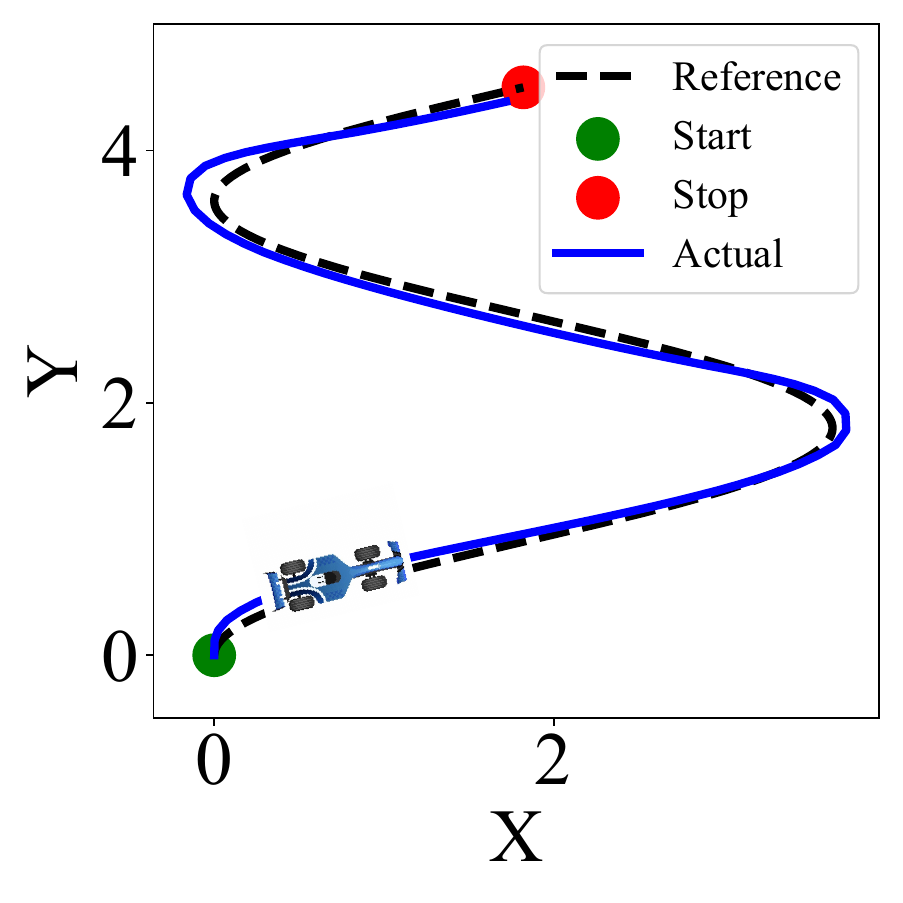}
    \caption{Model system}
    \label{subfig:f1tenth_Sim}
\end{subfigure}
\hspace{20pt}
\begin{subfigure}{0.4\linewidth}
    \includegraphics[height=1.3 in]{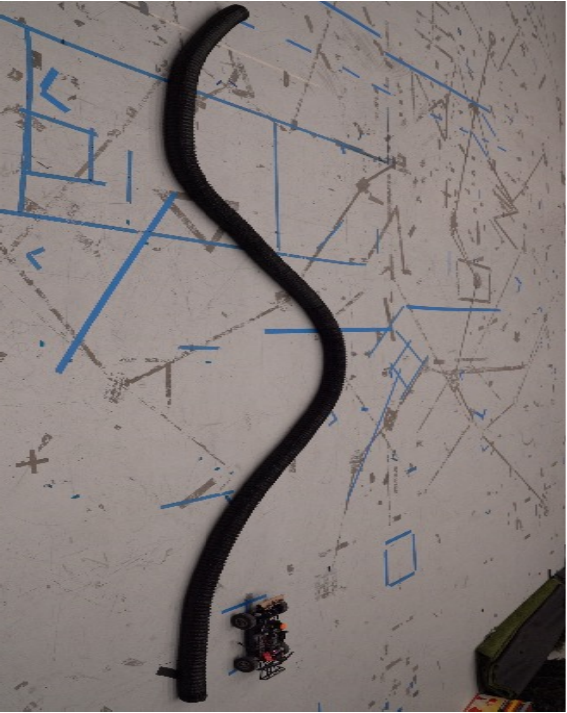}
    \caption{True system}
    \label{subfig:f1tenth_real}
\end{subfigure}
\caption{The simulation (model dynamics) environment and the hardware (actual dynamics) setup for the Push-T (top figures) and F1-Tenth (bottom figures) examples.  
 }
     \label{fig:pusht_setup}
\end{figure}

\section{Validations}\label{sec:validations}

\subsection{Diffusion Policy on the Push-T setup}\label{sec:pusht}

\textbf{Problem Description}: ~
In this example, we consider the task of pushing a T-block to a fixed target region using a manipulator arm equipped with a circular end effector. The control policy used for the task is a diffusion policy that uses expert demonstrations to learn the score function of a diffusion model for predicting actions conditioned on observations \cite{chi2023diffusionpolicy}.  The inputs generated by the policy $\pi$ consist of $(x_e,y_e)$, which corresponds to the  $(x, y)$ coordinates of the goal position of the circular end-effector. The policy is known to be robust to visual perturbations, and the implemented policy is trained in simulation using a \textsf{PyMunk} and \textsf{Gym} environment~\cite{florence2022implicit}, that simulates the contact dynamics of the T-block and the end-effector in a 2D plane. The model dynamics, in this case, is non-differentiable, represents the interactions of the end-effector and the T-block, and that of the T-block and the table, and does not take into account the kinematics and dynamics of the manipulator. Fig.~\ref{subfig:Pusht_Sim}-Fig. \ref{subfig:Pusht_real} show the simulation environment and the corresponding hardware setup, respectively. The environment variable $z$ was chosen as the initial position of the T-block ,i.e., $z = [x_b,y_b]$. The initial orientation of the block is kept fixed across all demonstrations to be the same as the target orientation of the block as shown in Fig. \ref{subfig:Pusht_real}.

\begin{figure}
    \centering
    \includegraphics[width=0.45\linewidth]{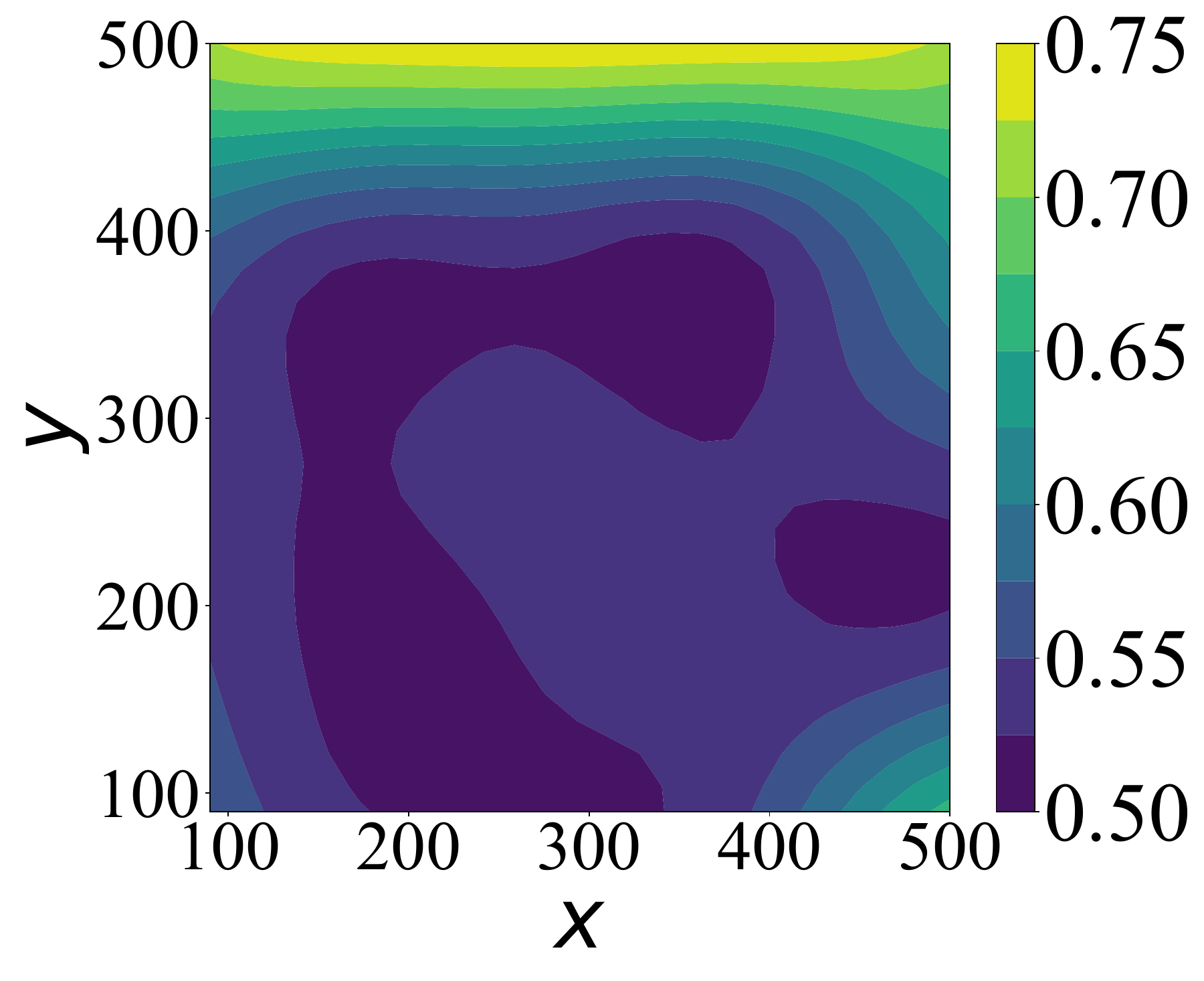}
    \includegraphics[width=0.45\linewidth]{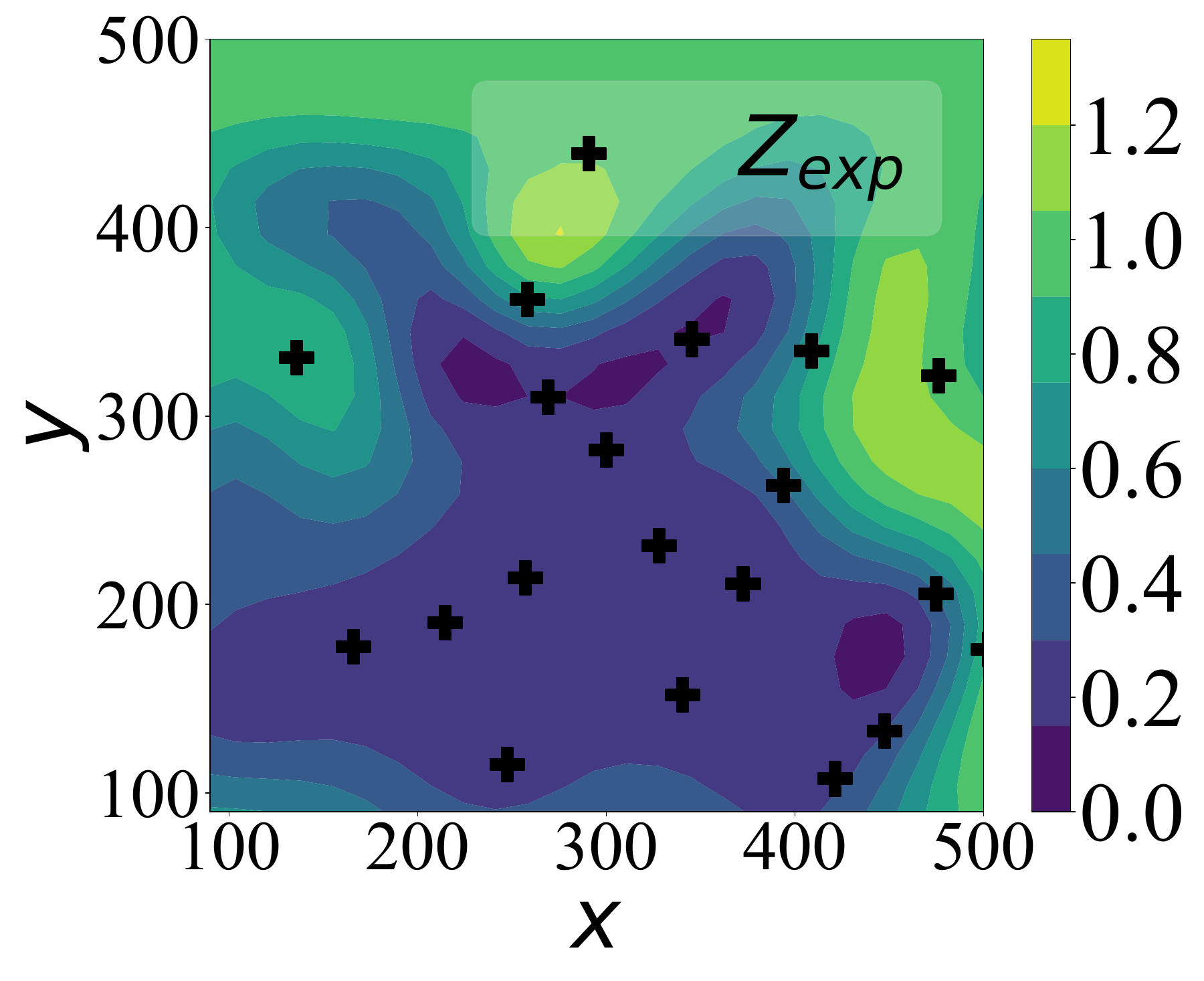}
    \caption{Risk predictions for Push-T example from GPR trained only simulation data (left) and using the proposed methodology (right). }
    \label{fig:Risk-contours}
\end{figure}

\textbf{Experimental setup}:~We use a UR3E Collaborative Robot Arm equipped with a Robotiq gripper to hold a 3D printed cylinder for the circular end-effector and a 3D printed T-block to construct the true system for the Push-T example. The diffusion policy in \cite{chi2023diffusionpolicy} is trained in simulation environment using pre-available dataset, where the workspace of the actual robot is not taken into consideration, and the end-effector is assumed to have only 2D motion in the 2D $xy$-plane. We used the Move-It package \cite{chitta2012moveit} to send commands to the manipulator for moving the end-effector to the desired goal location generated by the policy. 

For this problem, we set the number of demonstrations to $N=20$, with $N_1=10$ initial demonstrations. For discovering $\mathcal{Z}_{\text{risk}}$ from model dynamics, we chose $900$ uniformly spaced samples of $z=[x_b,y_b]$ from the domain $[100, 500]\times [100, 500]$ for each value of $2 \times 10^{-5}\leq \sigma_1 \leq 2\times 10^{-2}, 7\times 10^{-6}\leq \sigma_2 \leq  7\times10^{-3}$. Fig.~\ref{fig:method} (ii) shows the region $\mathcal{Z}_{\text{risk}}$ discovered in simulation using the model dynamics. The policy is trained using a reward $\gamma \in [0,1]$ that measures the maximum amount of overlap between the fixed target region and the  location of the T-block across a given rollout. We used $R = 1-\gamma$ as the risk function, where $R \geq 0.3$ (or equivalently, $\gamma \leq 0.7$) is obtained when the T-block does not have any overlap with the target region, and hence $R_{\text{th}}=0.3$ was set as the threshold for failure. The coverage function $C$ was chosen as the variance over $[X_b^1,\dots,X_b^T]$ given as $C(z,X_z) = \frac{1}{N}\sum_{t=1}^T\left(X_b^t - \sum_{t=1}^T\frac{X_b^t}{N}\right)^2$, where $X_b^t = [x_b^t,y^t_b]$ is the position of the T-block a time $0<t\leq T$, normalized to be within $[0,1]\times[0,1]$. This enables selecting values of $z$ that allow us to access a large part of the state space for each trajectory rollout, and thereby allow us to discover failures that are unseen in model dynamics. The coverage threshold was chosen as $C_{\text{th}}=0.15$ based on the coverage values from a few exemplar rollouts. 




\textbf{Result analysis}:~
Fig.~\ref{fig:Risk-contours} shows the contour for risk prediction using GPR with the proposed method labeled Sim+Exp (left) and using data collected with the model system only (right). The primary sources of sim-to-real gap in the Push-T example is the workspace constraints of the UR3E robot, which give rise to additional failures. This is not taken into account in the simulation environment and we discover these failures through the sequential demonstrations conducted using our approach in the paper, which leads to the difference in risk predictions using our method and model system only, as seen in Fig.~\ref{fig:Risk-contours}. 

\begin{figure}
    \centering
    \includegraphics[width=\linewidth]{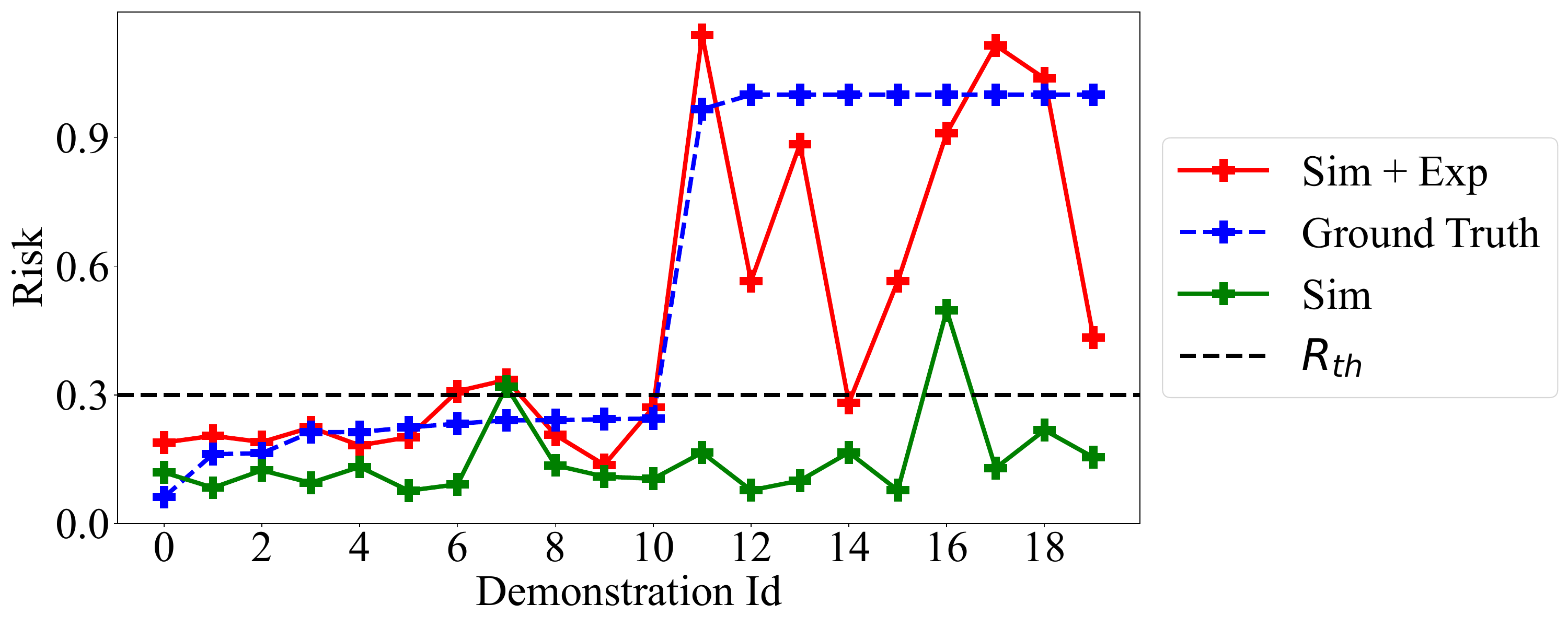}
    \caption{Model prediction for Push-T on $20$ data points where the predicted risk by the learned model $\phi_\theta$. The demonstration on true system marked as `Ground Truth' illustrates the prediction to be accurate. }
    \label{fig:pusht_figs}
\end{figure}

For validating the learned failure prediction using the learned model $\phi_\theta$, we record the risk for $20$ randomly sampled test demonstrations on the the true system, and compare against the predictions from our method. Table.~\ref{tab:PushT-Baselines} (Push-T) shows  the risk prediction error with GPR using two methods, namely, data collected only using model dynamics (reported as Simulation) and data collected using our approach (reported as Simulation+Exp). Fig. \ref{fig:pusht_figs} shows the individual predictions from our method (red), using data from model only (green) and their comparison against the ground truth (risk from true system, blue).  

\begin{table}[h]
\centering
\caption{Comparison of Sim and Sim+Exp risk predictors.}
\label{tab:PushT-Baselines}
\resizebox{0.8\linewidth}{!}
{%
\begin{tabular}{l c c c}
Env                       & Mode     & Sim   & Sim + Exp \\ \hhline{= = = =}
\multirow{2}{*}{Push-T}   & Fail     & 11\%  & 89\%      \\  
                          & Not Fail & 91\%  & 82\%      \\ \hline
\multirow{2}{*}{F1-Tenth} & Fail     & 36\%  & 100\%     \\
                          & Not Fail & 100\% & 100\%    
\end{tabular}
}
\end{table}

\subsection{LQR Speed+Steering control on the F1-Tenth car}\label{sec:F1}

\textbf{Problem Description}:~
In this example, we consider the problem of the F1-Tenth car tracking a reference race-line using a speed and steering LQR control. The policy is designed using a bicycle dynamic model with steering angle $\delta$ and acceleration $a$ as inputs, and input constraints $\lvert \delta \rvert \leq 0.8 \;\text{rad}$,  $\lvert a \rvert \leq 2 ms^{-2}$. The environment variable in this case is a combination of a parameter for the reference path to be tracked, which is set to be a sinusoidal curve with variable width $w$, and reference speed at which the vehicle is supposed to move, $v_{\text{ref}}$, i.e., $z=[w,v_{\text{ref}}]$. We implement the policy on the F1-Tenth platform which is considered as the true system.  Fig.~\ref{subfig:f1tenth_Sim} and Fig.~\ref{subfig:f1tenth_real} show an example reference trajectory being tracked by the model dynamics, and the corresponding hardware setup with the F1-Tenth vehicle denoting the true system, respectively. 

\begin{figure}
    \centering
    \includegraphics[width=0.435\linewidth]{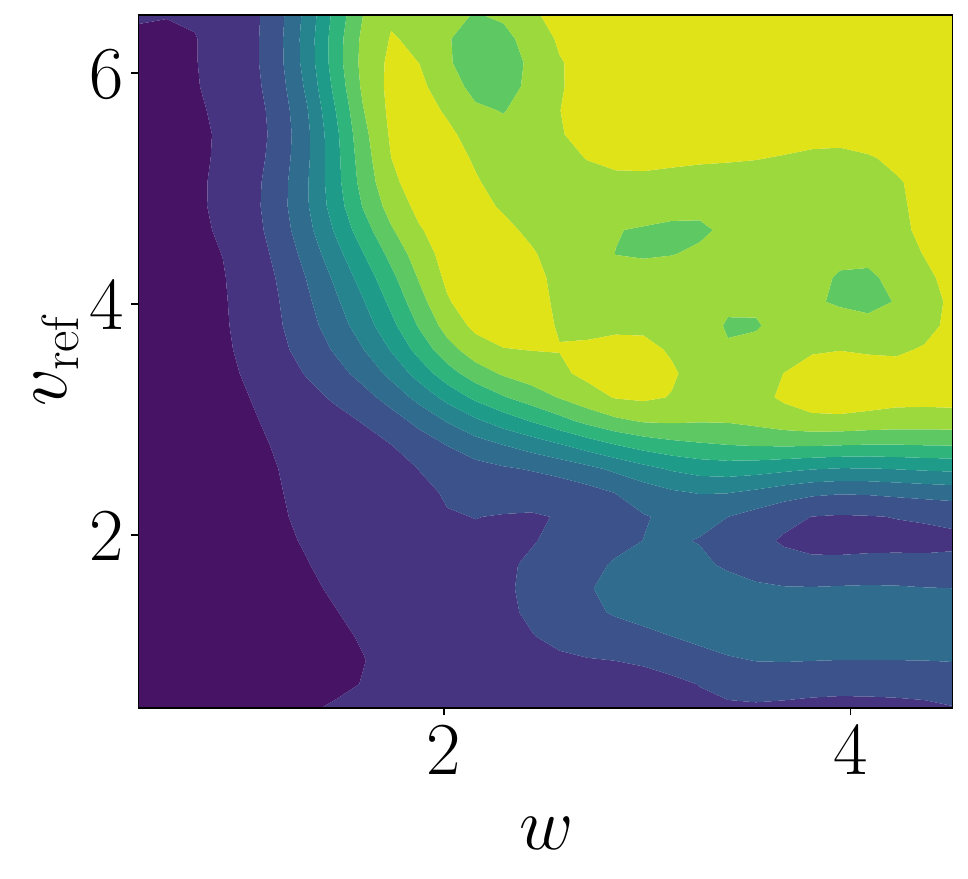}
    \includegraphics[width=0.455\linewidth]{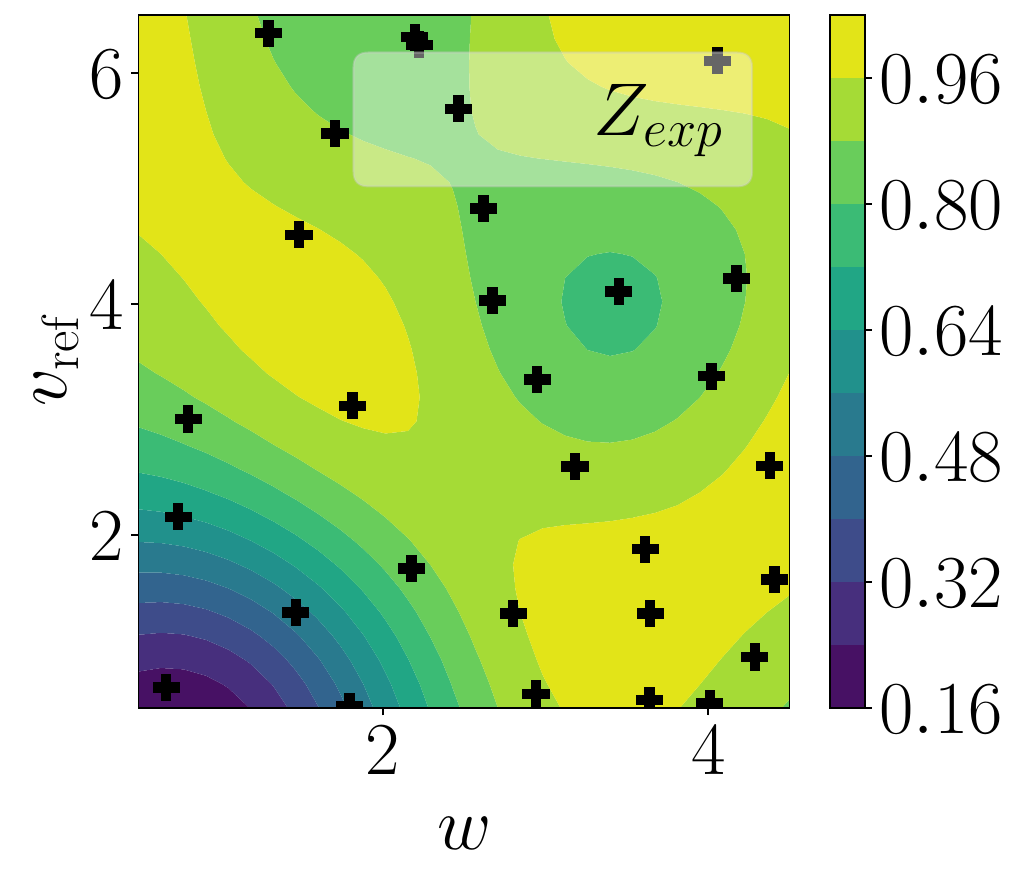}
    \caption{Risk predictions for F1-Tenth example from GPR trained only simulation data (left) and using the proposed methodology (right). }
    \label{fig:f1tenth gpr}
\end{figure}

\textbf{Experimental Setup:} The F1-Tenth vehicle used as the true system for hardware validation of the policy is known to have significant slip at high speeds $v\geq v_{\text{th}}$, and the proposed models for the vehicle in the literature consider a hybrid model to describe the lateral dynamics at high speeds, where the lateral dynamics plays an important role~\cite{althoff2017commonroad}. The performance of the vehicle at lower speeds is observed to be comparable to the bicycle model, however, the speed threshold $v_{\text{th}}$ at which the effect of slip becomes significant is unknown for the vehicle. To quantify this effect and uncover other possible causes of failures, we test across a diverse set of reference paths, where the width of the path leads to variation in steering angle requirement, in addition to the variation in the reference speed. 

We set the number of demonstrations to $N=30$, with $N_1=20$ initial demonstrations. For pre-processing failure information from model dynamics, $900$ uniformly spaced samples of $(w, v_\text{ref})$ are chosen from the domain $[0.5, 4.5]\times [0.5, 6.5]$
for each value of $10^{-4}\leq \sigma_1 \leq 1, 10^{-5}\leq \sigma_2 \leq 10^{-3}$. We choose a handcrafted risk function for this problem, consisting of the mean error ($R_{\text{mean}}$) and maximum error ($R_{\text{max}}$) between reference path and generated trajectory, and the minimum distance from the goal ($R_{\text{final}}$):
\begin{equation}\label{eq:risk_f110}
    R(z) = 20R_{\text{mean}}(z) + R_{\text{max}}(z) + 10 R_{\text{final}}(z)
\end{equation}
A failure in this context is defined as $R\geq R_{\text{th}}$, with $R_{\text{th}}=11.5$. The weight of each of the component functions and the threshold is chosen based on the threshold values for the individual terms that led to failures. For subsequent training, the risk is normalized using a \texttt{sigmoid} such that the new threshold is $R_{\text{th}}=0.5$. The coverage function $C$ is chosen as the range of the steering angle across a given trajectory rollout, because presence of steering angle limit is observed to be a dominant cause of failure in model dynamics, and a larger steering angle is required as the curvature of the race-line and reference speed increase. Therefore, for steering angle at time $t$ given by $\delta_t$, the coverage function can be expressed as $C(z,X_z) = \max\limits_{t\in [1,T]} \delta_t - \min\limits_{t\in [1,T]} \delta_t$. The coverage threshold in this case was chosen as $C_{\text{th}}=1.0 \; \text{rad}$, which is marginally above the steering angle limit.

\textbf{Result analysis}:~
Fig.~\ref{fig:f1tenth gpr} (right) shows the predicted risk contours using the trained model with the proposed method (Sim + Exp), while Fig.~\ref{fig:f1tenth gpr} (left) shows the predicted risk with a GPR model trained with data collected only from the model dynamics (Sim). Both the predictions show failure for high speed and high lane width, as the required steering angle increases as the speed or lane width increases. However, the threshold speed at which failure is observed in the true system is much lower, because the policy generates inputs without taking the lateral dynamics into account, which becomes increasingly prominent as the speed increases. This causes increasing tracking error, coupled with increasing steering angle requirement causes the failure region to expand substantially in the true system.  

\begin{figure}
    \centering
    \includegraphics[width=1\linewidth]{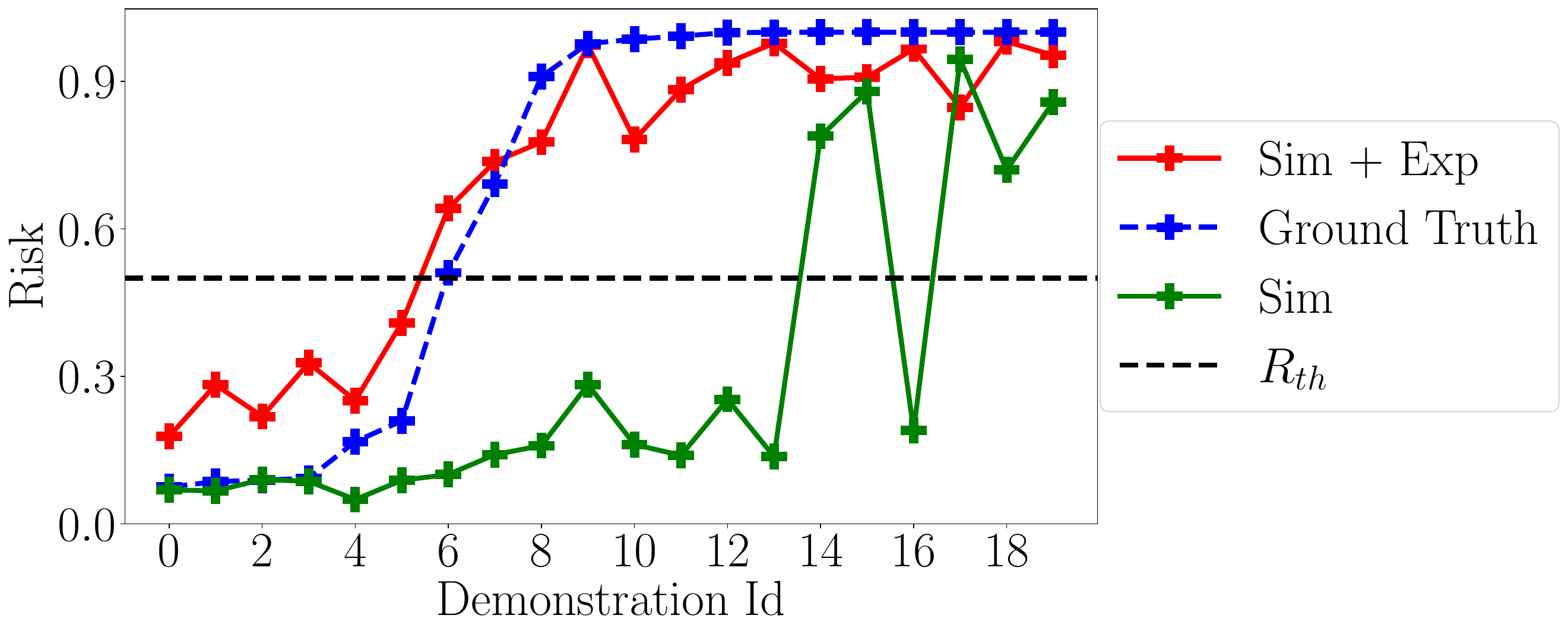}
    \caption{Model prediction for F1-Tenth on $20$ data points where the predicted risk by the learned model $\phi_\theta$. The demonstration on true system marked as `Ground Truth' illustrates the prediction to be accurate.}
    \vspace{-10pt}
    \label{fig:f1tenth pred error}
\end{figure}

Fig.~\ref{fig:f1tenth pred error} shows the predicted risk using the proposed method (red) and using just the model dynamics (green), compared against the ground truth risk (calculated using \eqref{eq:risk_f110}) obtained from the true system for $20$ randomly chosen $z$. Our method accurately predicts `fail' and `not fail' in all the $20$ cases. While the model trained only using model dynamics predicts the `not fail' scenarios accurately, it is unable to accurately forecast the `fail' scenarios correctly (see Table.~\ref{tab:PushT-Baselines} (F1-Tenth)), as it overestimates the `not fail' region.  The video corresponding to all hardware experiments can be found at the project website\footnote{https://mit-realm.github.io/few-demo/} 

\section{Conclusions \& Future Work}
\label{sec:conclusion}
In this work, we considered the problem of discovering failures in a system that emerge due to the sim-to-real gap. As a solution, we propose to use the available data from the simulation, which we refer to as the model dynamics, in addition to limited demonstrations from the true system. This allows us to leverage the benefits of simulation-based testing and hardware testing, while also maintaining the budget for hardware experimentation and overcoming the limitations arising from the lack of information. Our proposed method successfully predicts failures that cannot be captured by model dynamics alone, as illustrated on two different platforms, namely a UR3E manipulator arm and the F1-Tenth autonomous racing platform. Our experiments quantify the sim-to-real gap in the considered examples where the simulation-based failure predictor has a far less accuracy ($11\%$ and $36\%$ in the two examples) than the one obtained from the proposed method ($89\%$ and $100\%$, respectively).  

The proposed method deals with failure defined on the exogenous system parameters. In our future work, we will explore more general failure representation that are not scenario specific, e.g., set of failures defined on the state and input space. We are also interested in exploring how to integrate the proposed failure prediction methodology with existing risk-aware decision-making pipelines, that require knowledge of failure scenarios to prevent them. 


%


\newpage
\clearpage

\bibliographystyle{IEEEtran}
\bibliography{IEEEabrv,main}

\end{document}